\documentclass{svproc}
\usepackage{url}
\usepackage{amsmath}
\usepackage{amssymb}
\usepackage{graphicx}
\usepackage{subcaption}
\usepackage{xcolor}
\usepackage{array}
\newcommand{\PreserveBackslash}[1]{\let\temp=\\#1\let\\=\temp}
\usepackage{multirow}
\newcolumntype{C}[1]{>{\PreserveBackslash\centering}p{#1}}
\newcolumntype{R}[1]{>{\PreserveBackslash\raggedleft}p{#1}}
\newcolumntype{L}[1]{>{\PreserveBackslash\raggedright}p{#1}}
\begin{document}
\mainmatter              % start of a contribution
\title{FAMSeg: Fetal Femur and Cranial Ultrasound Segmentation Using Feature-Aware Attention and Mamba Enhancement}
\titlerunning{FAMSeg Net}  % abbreviated title (for running head)
\author{Jie He\inst{1} \and Minglang Chen\inst{2,1,*} \and
Minying Lu\inst{3} \and Bocheng Liang\inst{4} \and Junming Wei\inst{6} \and Guiyan Peng\inst{4} \and Jiaxi Chen\inst{5}\and Ying Tan\inst{4}}
\authorrunning{Jie He et al.} % abbreviated author list (for running head)
% \tocauthor{Ivar Ekeland, Roger Temam, Jeffrey Dean, David Grove,
% Craig Chambers, Kim B. Bruce, and Elisa Bertino}
%
\institute{Guangxi Key Laboratory of Machine Vision and Intelligent Control, Wuzhou University, Wuzhou, 543002 China. \and Faculty of Innovation Engineering, Macau University of Science and Technology, Macau, 999078 China.\\
\email{2240030650@student.must.edu.mo.}%,\\ WWW home page:
% \texttt{http://users/\homedir iekeland/web/welcome.html}
\and School of Computer Science and Information Security, Guilin University of Electronic
Technology, Guilin, 541004 China.\and 
Shenzhen Maternity and Child Healthcare Hospital, Southern Medical University, Shenzhen, 518100 China. \and College of Big Data and Software Engineering, Wuzhou University, Wuzhou, 543002 China. \and College of Electronical and Information Engineering Wuzhou University, Wuzhou,543002 China.}
\maketitle              % typeset the title of the contribution
\vspace{-10mm}
\begin{abstract}  
Accurate ultrasound image segmentation is a prerequisite for precise biometrics and accurate assessment. Relying on manual delineation introduces significant errors and is time-consuming. However, existing segmentation models are designed based on objects in natural scenes, making them difficult to adapt to ultrasound objects with high noise and high similarity. This is particularly evident in small object segmentation, where a pronounced jagged effect occurs. Therefore, this paper proposes a fetal femur and cranial ultrasound image segmentation model based on feature perception and Mamba enhancement to address these challenges. Specifically, a longitudinal and transverse independent viewpoint scanning convolution block and a feature perception module were designed to enhance the ability to capture local detail information and improve the fusion of contextual information. Combined with the Mamba-optimized residual structure, this design suppresses the interference of raw noise and enhances local multi-dimensional scanning. The system builds global information and local feature dependencies, and is trained with a combination of different optimizers to achieve the optimal solution. After extensive experimental validation, the FAMSeg network achieved the fastest loss reduction and the best segmentation performance across images of varying sizes and orientations.
\end{abstract}
\section{Introduction}
Accurate delineation of key anatomical structures provides detailed spatial localization and applicable measurement methods for structural biometrics. Precise biometric measurements of fetal ultrasound images are essential for accurately assessing fetal health and development~\cite{pu2021automatic}. However, in clinical practice, delineating key anatomical structures often relies on manual tracing by ultrasound technicians. This not only increases the workload but also~\cite{gao2024graph}, due to the reliance on visual inspection and physiological limitations, often results in inaccurate or irregular boundary delineation~\cite{pu2024unsupervised}. Therefore, by integrating deep learning methods, key anatomical structures can be accurately delineated at the pixel level, efficiently assisting ultrasound technicians. This reduces their workload, improves boundary delineation accuracy, and minimizes measurement errors, ultimately enhancing prediction and assessment accuracy.

Fetal ultrasound imaging is influenced by multiple factors~\cite{zhao2024farn}, including amniotic fluid, acquisition parameters~\cite{zhao2022ultrasound}, and scanning techniques~\cite{pu2024m3}, leading to issues such as small anatomical structures, significant noise interference, and low contrast. Existing deep learning methods struggle to capture subtle variations in structures like the fetal femur and cranial region. However, most methods aimed at achieving efficient and accurate segmentation employ techniques such as fine-tuning the encoder, introducing attention mechanisms, and deepening the decoder to obtain richer feature maps that complement the encoder’s texture extraction. As the feature map dimensions increase, the receptive field becomes larger, which, while improving global context, reduces the ability to accurately segment small targets and define boundaries~\cite{yang2025bidirectional}, leading to more pronounced jagged edges in the segmentation results.

Thus, this study proposes an end-to-end segmentation framework based on the Mamba mechanism and adaptive feature-awareness. By designing various convolution methods to more accurately capture contour information and integrating adaptive feature-aware enhancement for local dependencies, the approach addresses key challenges in fetal femur and cranial ultrasound segmentation. Our method offers the following contributions.
\begin{itemize}
\item {Design a multi-branch deep strip convolution module to independently scan feature maps from both horizontal and vertical perspectives, enhancing the description of target boundary features.}
\item {Design a Mamba residual model to reduce the propagation of original features in the network through Mamba’s multi-view scanning approach, minimizing interference with segmentation accuracy.}
\item {Design an adaptive perception module to capture the weight information of different features in the feature map, applying weighted adjustment to the original features to enhance the model's nonlinear mapping capability.}
\item {Design a multi-optimizer alternating scheme to address issues of the network missing the optimal solution and failing to reach convergence.}
\end{itemize}

\section{Related Work}
With the development of supercomputers with powerful computational capabilities, deep convolutional neural networks (CNNs) with large-scale parameters and complex computational structures have achieved significant success in image processing and medical image analysis and diagnosis. In 2024, Chen et al.~\cite{CHEN2024107898} proposed a feature encoding module for multi-view local perspectives, designed multi-channel capture to model global dependencies between different nodes, and constructed semantic relationships between features at different levels to enhance the extraction and fusion of global and local features, thereby improving the model's segmentation performance. In 2022, Lu et al.~\cite{9950712} introduced a novel multi-level non-maximum suppression (NMS) mechanism to further enhance the segmentation performance across three selection levels and performed instance segmentation of 13 anatomical structures in fetal four-chamber ultrasound images. In 2022, Pu et al.~\cite{pu2022mobileunet} proposed the construction of an explicit FPN network to enhance multi-scale semantic information fusion, improving MobileNet's performance in the apical four-chamber segmentation of fetal echocardiograms.

Due to the limited availability of open-source medical image data, most semantic segmentation models are designed for industrial or natural objects and often struggle with medical image analysis. Consequently, many researchers enhance feature extraction by adopting semi-supervised networks or introducing attention mechanisms. In 2024, Chen et al.~\cite{chen2024sam2} proposed an adaptation module for fine-tuning downstream tasks under the SAM2 large model, improving complex segmentation tasks such as camouflage. In 2024, Xing et al.~\cite{xing2024segmamba} integrated the Mamba mechanism to capture long-range dependencies within volumetric features at various scales, enhancing the network's segmentation performance.

This paper enhances contour and small target information extraction by designing distinct perspective scanning methods, reducing model complexity. By integrating the Mamba V2~\cite{mamba2} mechanism for multi-view scanning, it captures both local and global feature dependencies. An adaptive perception module and original feature fusion method are employed to improve the decoder's nonlinear mapping capability. Finally, alternating optimization with different optimizers is used to limit model fluctuations around the optimal solution, thus improving segmentation performance.

\section{Method}
\subsection{FAMSeg overview}
\begin{figure}[htbp]
\centerline{\includegraphics[width=\textwidth]{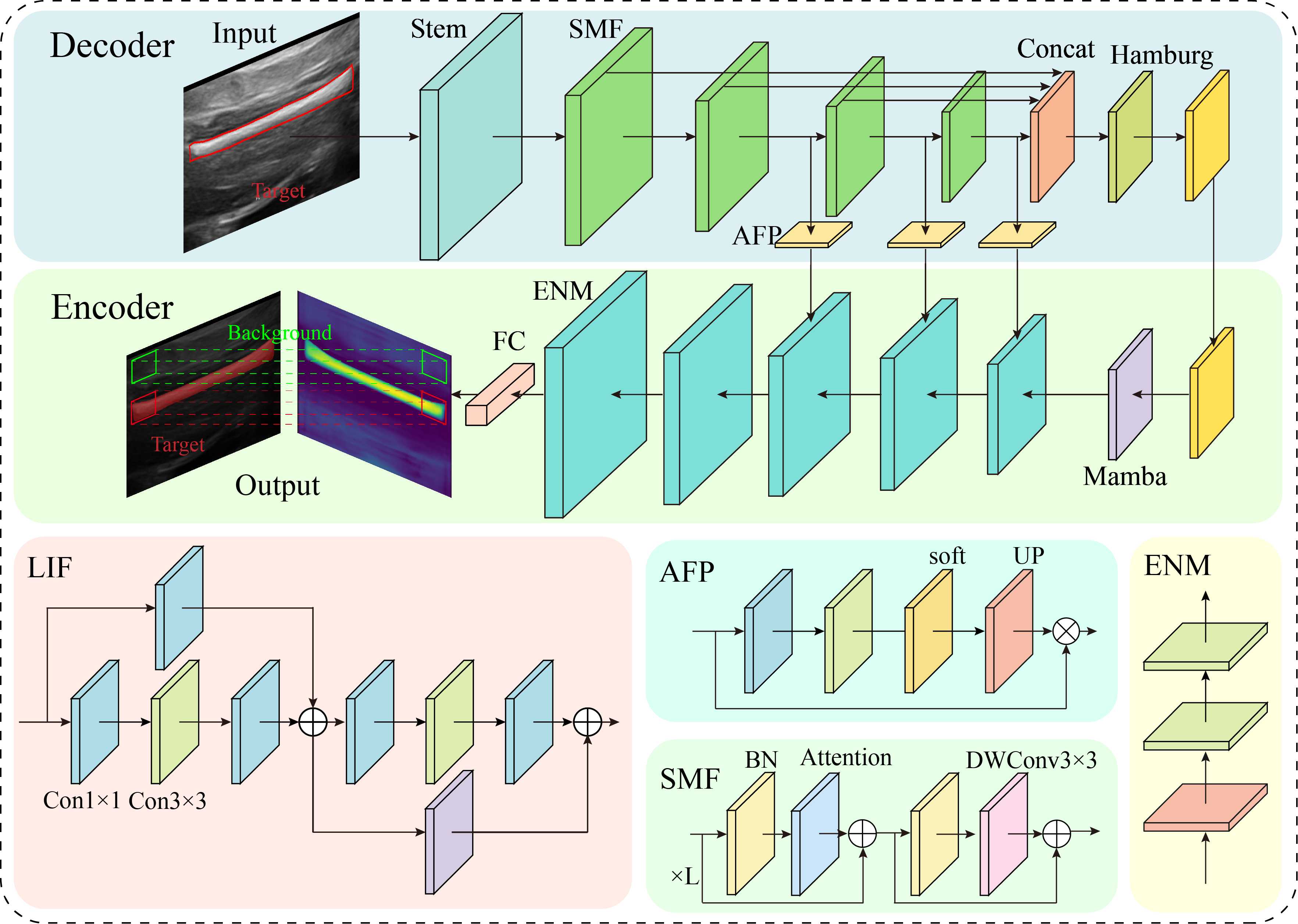}}
\caption{The schematic diagram of the model details for FAMSeg.}
\label{fig1}
\end{figure}

Fig.~\ref{fig1}, illustrates the framework of the proposed fetal femur and cranial ultrasound image segmentation model. The shallow features have a small receptive field and exhibit strong capabilities in capturing edge and texture information. However, to balance the segmentation of large and small targets, many models typically construct a simple low-level feature extraction module and a complex high-level feature extraction module. The fusion of high-level and low-level features enhances the segmentation performance for small targets. This approach, which supplements high-level features while neglecting low-level feature extraction, performs well for object segmentation in natural scenes. However, it struggles to adapt to the segmentation of objects in medical ultrasound images, which often exhibit highly similar, blurred, and low-contrast features. To address the limitation of the SegNeXt~\cite{segnext} model in segmenting small objects, we introduce the Mamba mechanism and design a context feature fusion module. This enhances the communication of local feature information and the integration of multi-dimensional data within the shallow network, thereby improving the segmentation accuracy of small structures in medical images.
\vspace{-4mm}
\vspace{-13mm}
\subsection{Feature Encoder Design}
Most encoders are primarily composed of convolutional modules or residual structures in a unidirectional concatenation manner. This results in weak local feature flow and extraction, making it difficult to deeply mine and express key object edge contour information. Consequently, the decoder lacks edge reference during feature recovery, leading to pronounced aliasing effects and the disappearance of small object features. To address the issues of aliasing and small object feature disappearance, the encoder in this work is primarily composed of multi-branch deep strip convolution modules and the Mamba residual structure, effectively integrating object texture information.

\subsubsection{Multi-branch deep strip convolution}
\begin{figure}[htbp]
    \centering
    \begin{subfigure}{0.45\textwidth}
        \centering
        \includegraphics[width=\textwidth]{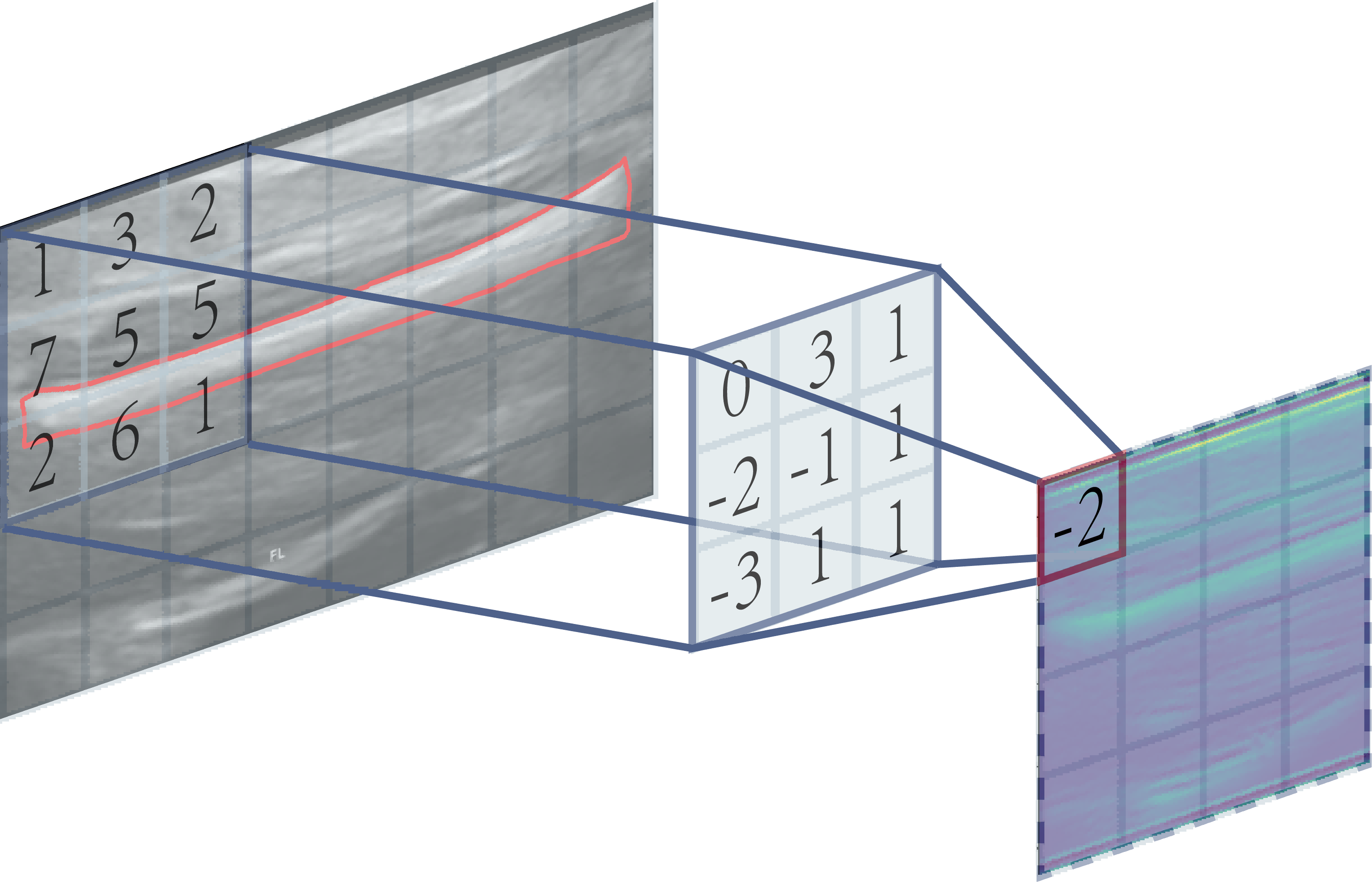}
        %\captionsetup{labelformat=empty} % 去掉子图编号
        \caption{Convlution block}
    \end{subfigure}%
    \hfill
    \begin{subfigure}{0.45\textwidth}
        \centering
        \includegraphics[width=\textwidth]{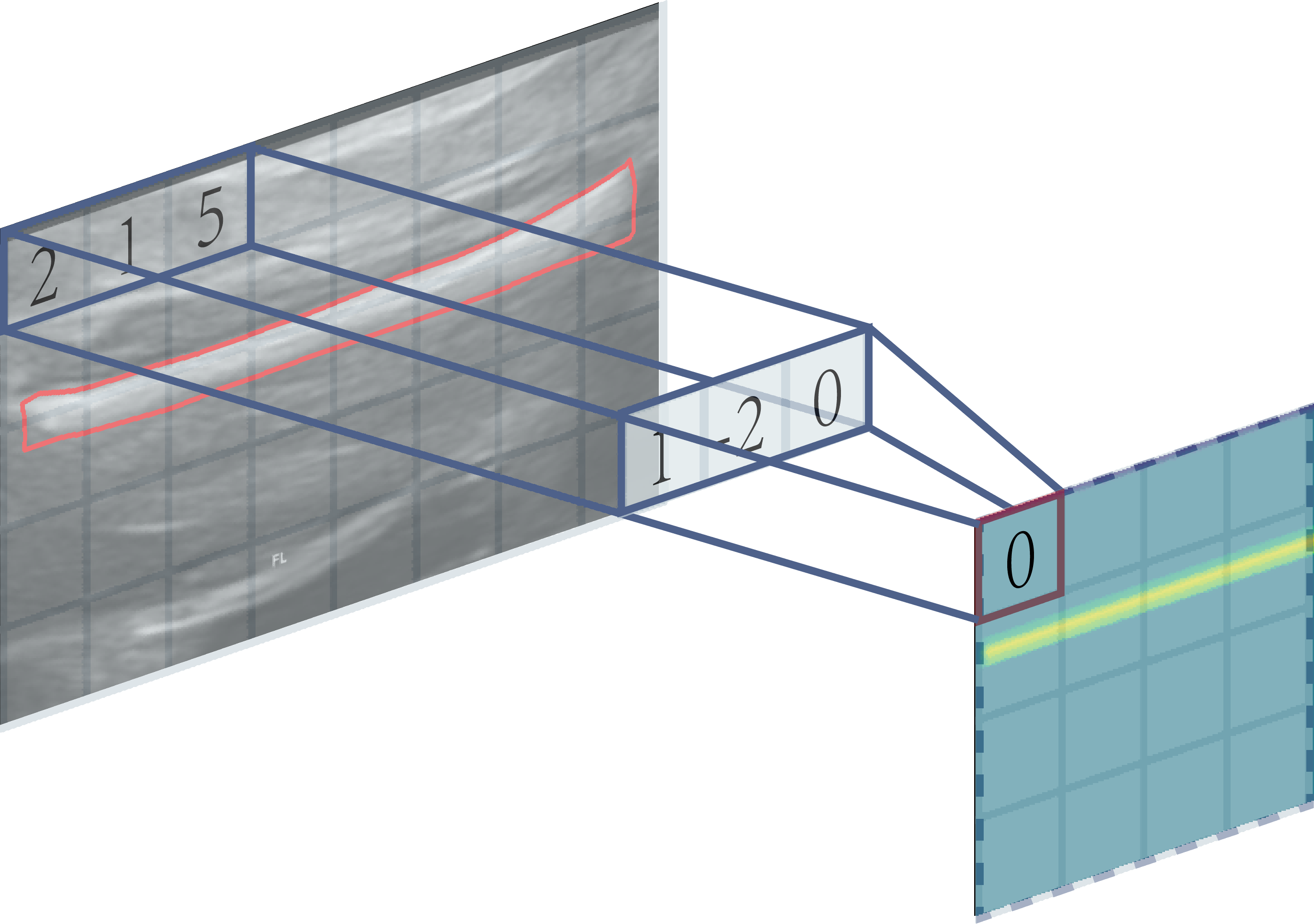}
        %\captionsetup{labelformat=empty} % 去掉子图编号
        \caption{Strided Convolution Module}
    \end{subfigure}
    \caption{The Principles of Different Convolution Operations. Scanning features from both horizontal and vertical independent perspectives are more effective in preserving object edge information.}
    \label{figxs}
\end{figure}

The multi-branch deep strip convolution method performs focused scanning of low-level features from both horizontal and vertical perspectives in a more computationally efficient manner. This reduces the computational cycle and mitigates feature degradation when scanning features from both directions simultaneously. As shown in Fig.~\ref{figxs} (a), standard convolution scans features both horizontally and vertically using a \(K\times K\) sliding window formed by the convolution kernel K. The window slides across the input feature map with a specific stride S. For each covered feature region, the elements of the convolution kernel are multiplied by the corresponding elements of the input feature map, and the results are summed to complete the feature extraction. Standard convolution captures abstract and rich feature information through simultaneous horizontal and vertical scanning. However, the dual-view approach tends to overlook smaller edge details, leading to the loss of small object information and aliasing issues during feature recovery. Therefore, as shown in Fig.~\ref{figxs} (b), we design a multi-branch deep strip convolution module that employs multiple convolution kernels and unidirectional scanning to supplement the missing edge and contour information in low-level abstract features.
\vspace{-11mm}
\vspace{-9.4mm}

First, abstract features are extracted using a \(5\times 5\) convolution. Then, multiple convolution kernels scan the features separately from horizontal and vertical perspectives. Finally, through feature stacking, unidirectional information is integrated into complex multi-view information, thereby alleviating the issues of small object and contour feature loss. Additionally, to enhance the receptive field of the encoder's unidirectional features, we design an aggregated depth convolution module to fuse feature information from different encoder layers. Feature enhancement is further achieved using the hamburger module from the SegNeXt network, thereby improving the model's segmentation performance and robustness. Moreover, the multi-branch deep strip convolution module offers higher computational efficiency compared to standard convolution, as shown in Eq.~(\ref{eq1}) and Eq.~(\ref{eq2}),
\begin{align} 
% Parameters_{conv}=(H-K+1)^2\times K^2
    Parameters_{conv}=K\times H_o \times W_o\times K ,
    \tag{1} 
    \label{eq1}
\end{align}
\vspace{-5mm}
    \begin{align} 
    Parameters_{our}=2\times H_o \times W_o\times K .
    \tag{2} 
    \label{eq2}
\end{align}

\vspace{-3mm}
Let \( H_i \) and \( W_i \) represent the input feature dimensions, and \( H_o \) and \( W_o \) the output feature dimensions. The stride is denoted as \( S \). When the kernel size \( K = 7 \), the computational cost for both horizontal and vertical directions is given by
\begin{align} 
    {Cost}_{7\times7} = 2 \times H_o \times W_o \times 7,
    \tag{3}
    \label{eq3}
\end{align}
\vspace{-3mm}
whereas for a standard convolution, the cost is
\begin{align}
    {Cost}_{7\times7\text{(standard)}} = 7 \times H_o \times W_o \times 7.
    \tag{4}
    \label{eq4}
\end{align}

If a \( 3 \times 3 \) convolution is used instead, with the same output, the computational cost becomes
\begin{align}
{Cost}_{3\times3} = 2 \times H_o \times W_o \times 9.
\tag{5}
\label{eq5}
\end{align}
\vspace{-3mm}

Therefore, it is evident that independent scanning leads to a lower computational cost.

\subsubsection{Mamba Residual Structure}
To achieve more stable model propagation and obtain precise segmentation masks, we introduce two residual structures: the Convolution block and the Bottle block. The Convolution block downsamples input features, reducing computational data and providing a richer receptive field for the Bottle block. This enables deeper mining and learning of abstract features, enhancing both local and global feature dependencies. It improves the receptive field of the multi-branch deep strip convolution module, addressing the limited feature diversity in single-view scanning and refining the model's segmentation of contours.

The main branches of both blocks consist of two \(1\times 1\) convolutions and one \(3\times 3\) convolution. Their key distinction lies in the side branches: the Convolution block’s side branch includes a \(1\times 1\) convolution for feature fusion post-downsampling, while the Bottle block’s side branch directly shortcuts to the output features for stacking and feature extraction. These residual structures, through information branching, help mitigate gradient vanishing.

However, shortcut connections can allow unfiltered noise to enter the network, especially when multiple Bottleneck modules are stacked. This leads to the accumulation of raw noise in output features, reducing the filtering capacity of subsequent layers and impacting network stability and segmentation accuracy. To address this, we optimize the Bottleneck shortcut branch by introducing the Mamba V2 module, which filters out noise and enhances low-level feature and contour extraction in the encoder through its multi-angle scanning approach.

\subsection{Feature-Aware Decoder}
Low-level abstract features provide the encoder with feature maps that capture global image information. However, these abstract feature maps lose local information and spatial details, making it impossible to directly generate segmentation masks from them. Relying on single-feature upsampling to gradually restore resolution often leads to the loss of target details and spatial information, especially in ultrasound image segmentation tasks where key anatomical structures are unclear and image quality is poor. However, the original features from the encoder contain high-resolution information, which is crucial for precise localization and edge segmentation. Therefore, the decoder combines feature fusion and upsampling methods to progressively refine abstract feature information. By integrating a feature-aware attention mechanism, it enhances the nonlinear capability of the features, restoring the spatial resolution of the original image to generate the segmentation mask.

Although the original features excel in precise localization and edge segmentation, they contain significant background noise. However, traditional feature upsampling methods typically perform simple spatial expansion on the input feature map, often leading to image blurring or loss of details. Therefore, the FAM mechanism introduces a dynamic content-aware approach, using the spatial distribution at each position of the input feature map as weights. This combines and extends different parts of the feature map with mapping constraints, making upsampling not just a simple spatial expansion, but a process that incorporates global information from the input features to generate finer, high-quality feature maps.
\vspace{-6.5mm}
\begin{figure}[htbp]
    \centerline{\includegraphics[width=\textwidth]{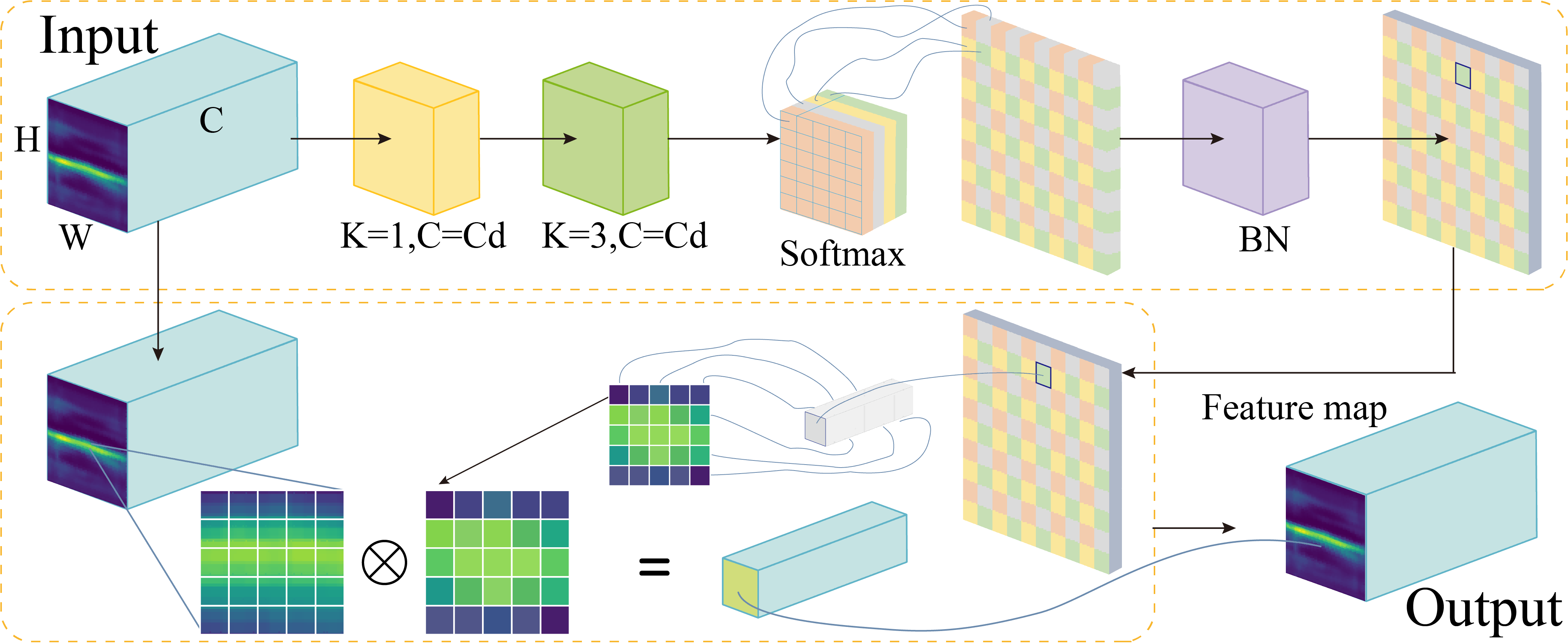}}
    \caption{Schematic diagram of the Feature-Aware Attention Mechanism principle.}
    \label{fig2}
\end{figure}
\vspace{-5mm}

As shown in Fig. ~\ref{fig2}, the FAM consists mainly of a channel compressor, kernel reorganizer, and kernel normalization. First, the \(1\times 1\) channel compressor reduces the input feature channels from C to Cd, thereby decreasing the computational complexity of the module by reducing the number of channels in the feature map. Next, a kernel reorganizer with a kernel size of k generates reorganized kernels based on the content of the channel-compressed feature map, forming a feature map of size \(K\times K \times C\). This expands the receptive field and enables feature nonlinear mapping using contextual information from a larger region. Then, the softmax function normalizes the features in terms of spatial information to assess their importance. Finally, the feature importance is weighted with the original features, dynamically perceiving and reorganizing them to enhance the model’s understanding.

To reduce interference from raw noise, we fuse only the features from the last three layers of the encoder. Additionally, we apply convolution functions to the upsampled and feature-fused feature maps to enhance nonlinear feature mapping and extract local features, improving the model’s ability to capture more complex features and further refine the feature map. Once the feature map is restored to the original resolution, a \(1\times 1\) convolution is used instead of a fully connected method to identify and classify the features in the feature map, completing the anatomical structure segmentation task.

\subsection{Optimizer and Loss Minimization Method}
The mainstream optimizers currently include Stochastic Gradient Descent (SGD)~\cite{SGD}, Adam with Weight Decay (AdamW)~\cite{adamw}, and Adaptive Moment Estimation (Adam)~\cite{adam}. However, each optimizer has its own significant advantages and inevitable drawbacks. Therefore, to achieve more efficient learning and higher segmentation accuracy, this paper uses both SGD and AdamW optimizers, alternating between them to leverage their respective advantages. This approach complements the limitations of each optimizer, reducing the learning cost and enhancing segmentation performance.

The SGD optimizer calculates the model gradient using only one sample at a time, making it easy to implement with minimal hyperparameters. Compared to some adaptive optimizers, it offers better generalization capability. SGD is known for its simplicity, computational efficiency, and strong generalization ability. As shown in Eq.~(\ref{eq6}), the SGD optimizer calculates without incorporating gradient momentum, and typically requires a learning rate decay strategy for optimal performance. Therefore, the SGD optimizer is typically slower and requires longer fitting cycles, but it can oscillate around the optimal solution, making it suitable for fine-tuning the model to reach the optimal solution.
\begin{align} 
    \theta_{t+1} = \theta_t - \eta \cdot \nabla_\theta L(\theta_t) .
    \tag{6} 
    \label{eq6}
\end{align}

However, both Adam and AdamW optimizers incorporate gradient momentum and variance to optimize the network, significantly improving optimization efficiency compared to SGD, enabling faster model convergence and reducing training time. Although the Adam optimizer is far more efficient than SGD, it is highly dependent on the initial learning rate. A large initial learning rate may cause the network to miss the optimal solution, while a small one may prevent convergence. However, the AdamW optimizer applies weight decay to the parameter updates instead of updating it alongside the gradient, which not only addresses the reliance on the initial learning rate but also provides more stable and improved optimization performance.
\begin{align}
    m_t = \beta_1 m_{t-1} + (1 - \beta_1) \nabla_\theta L(\theta_t), v_t = \beta_2 v_{t-1} + (1 - \beta_2) \nabla_\theta L(\theta_t)^2 ,
    \tag{7} 
    \label{eq7}
\end{align}
\vspace{-6mm}
\begin{align}
    \hat{m}_t = \frac{m_t}{1 - \beta_1^t}, \quad \hat{v}_t = \frac{v_t}{1 - \beta_2^t} ,
    \tag{8} 
    \label{eq8}
\end{align}
\vspace{-6mm}
\begin{align}
    \theta_{t+1} = \theta_t - \eta \cdot \frac{\hat{m}_t}{\sqrt{\hat{v}_t} + \epsilon} ,
    \tag{9} 
    \label{eq9}
\end{align}
\vspace{-6mm}
\begin{align}
    \theta_{t+1} = \theta_t - \eta \cdot \frac{\hat{m}_t}{\sqrt{\hat{v}_t} + \epsilon} - \eta \cdot \lambda \cdot \theta_t .
    \tag{10} 
    \label{eq10}
\end{align}

In summary, we first use the AdamW optimizer for 150 epochs to quickly converge near the optimal solution, avoiding under-iteration that may prevent reaching the optimal solution. Then, we switch to the SGD optimizer for 50 epochs to allow sufficient optimization, ensuring the network reaches the optimal gradient solution.

\section{Experiments and Visualization}
\subsection{Dataset}
We collected a dataset of 3,798 ultrasound images of fetal femur and cranial structures from Shenzhen Maternal and Child Health Hospital, covering different developmental stages. The fetal femur is abbreviated as FL, and the cranial structure as FB. The dataset was collected using equipment from Philips and Siemens during the 14 to 28 weeks of pregnancy. 
% Table X shows the random division of the dataset into training, validation, and test sets in a 7:2:1 ratio.
\subsection{ Experimental Platform}
All experiments were conducted on an Arch Linux system, equipped with an Intel Xeon Platinum 8360Y CPU and four NVIDIA A100 GPUs, using PyTorch as the framework for model development. The maximum and minimum learning rates during initialization are set to 0.01 and 0.0001, respectively, with the upper and lower bounds of the learning rate being 0.001 and 0.0001. To dynamically adjust the learning rate and enhance the model's stability, we adaptively adjust the maximum and minimum learning rates based on the batch size.
\begin{align}
    \text{Init\_lr\_fit} &= \min \left( \max \left( \frac{\text{batch\_size}}{64} \times \text{init\_lr}, \, \text{lr\_limit\_min} \right), \, \text{lr\_limit\_max} \right) ,
    \tag{11} 
    \label{eq11}
\end{align}
\vspace{-6mm}
\begin{align}
    \text{Min\_lr\_fit} &= \min \left( \max \left( \frac{\text{batch\_size}}{64} \times \text{min\_lr}, \, \frac{\text{lr\_limit\_min}}{100} \right), \, \frac{\text{lr\_limit\_max}}{100} \right) .
    \tag{12} 
    \label{eq12}
\end{align}
\subsection{Comparison of Segmentation Performance}
\begin{figure}[htbp]
\centerline{\includegraphics[width=\textwidth]{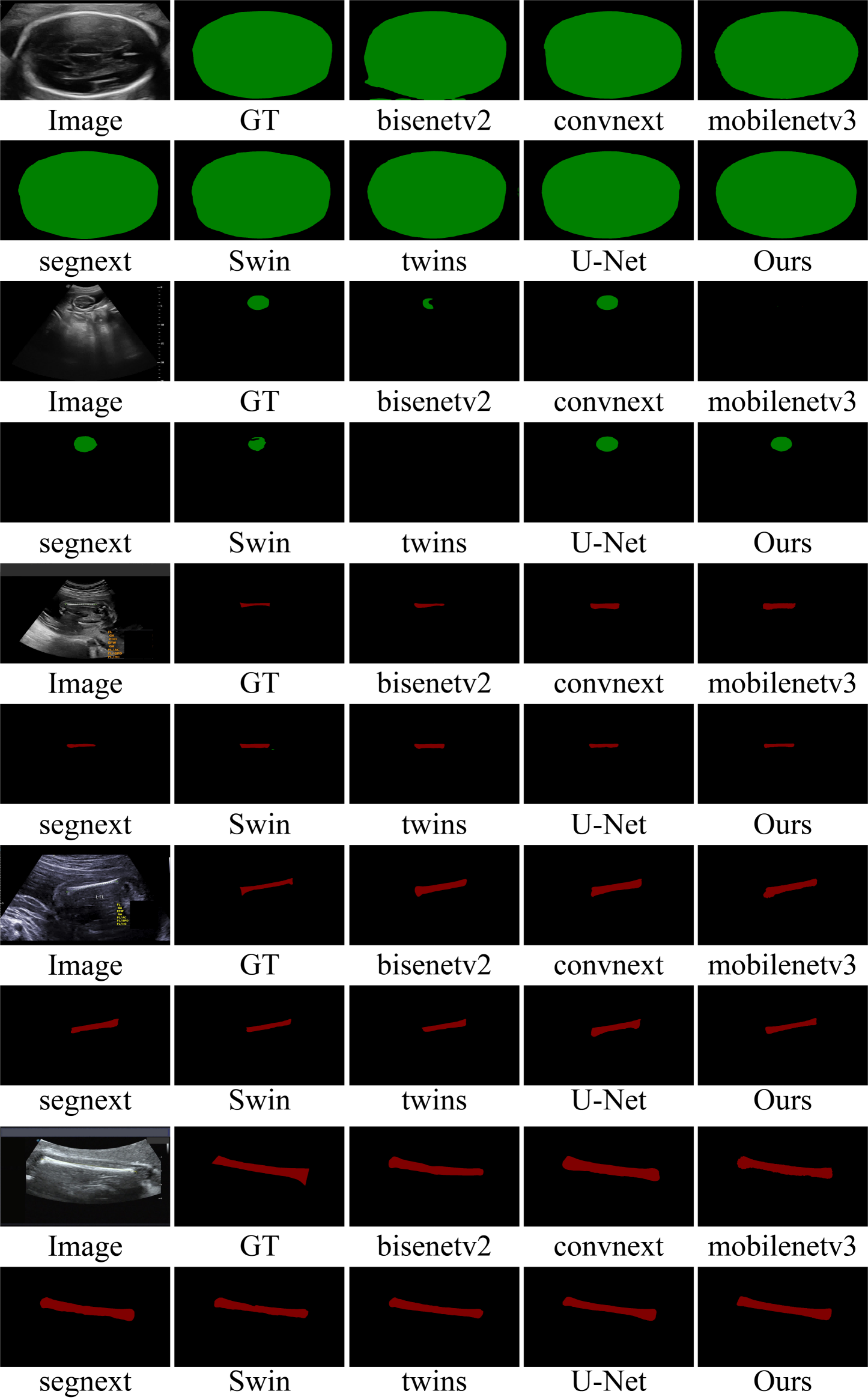}}
\caption{Visualization of the comparison and validation results of mainstream baseline models. The segmentation results show the brain and skull in green, and the femur in red.}
\label{fig3}
\end{figure}

To thoroughly evaluate the segmentation and generalization performance of the FAMSeg model, we design comparisons with mainstream or relatively recent segmentation models such as BisNet V2 and U-Net. To ensure consistent measurement, we use the IoU (Intersection over Union) parameter to evaluate the segmentation results. The experimental results are shown in Table \ref{tab2}. The FAMSeg model not only outperforms other networks but also achieves superior results on the femur dataset, which contains a significant number of small targets. The segmentation results of each model are shown in Fig.~\ref{fig3},
%\vspace{-9mm}
\begin{table*}
    \centering
    \setlength{\tabcolsep}{10pt}
    % \resizebox{\linewidth}{!}{
    \caption{Comparison and validation results of mainstream baseline models.}
    \begin{tabular}{l|cccc}
    \hline
        Item & BG & FL & FB & mIoU$\uparrow$  \\ \hline
        U-net\cite{U-Net} & 98.08  & \underline{78.30}  & 94.02  & 90.13   \\ %\hline
        twin\cite{chu2021twins}  & 98.95  & 76.43  & 94.43  & 89.94   \\ %\hline
        Swin~\cite{liu2021Swin} & \underline{99.05}  & 76.92  & \underline{95.05}  & \underline{90.34}   \\ %\hline
        SegNeXt & 98.16  & 77.48  & 94.18  & 89.94   \\ %\hline
        mobilenet v3~\cite{Howard_2019_ICCV} & 98.87  & 71.68  & 94.56  & 88.37   \\ %\hline
        ConvNeXt~\cite{liu2022convnet} & 98.84  & 71.30  & 94.94  & 88.36   \\ %\hline
        BisnetV2~\cite{yu2021bisenet} & 98.76  & 73.03  & 93.95  & 88.58   \\ %\hline
        Ours & \textbf{99.08}  & \textbf{80.21}  & \textbf{95.44}  & \textbf{91.58}  \\ \hline
    \end{tabular}
    \label{tab2}
    \vspace{-6.7mm}
\end{table*}

The FAMSeg model effectively performs ultrasound image segmentation of fetal bones and cranial structures with varying shapes, sizes, and orientations, without noticeable issues such as jagged effects, false positives, false negatives, or misclassifications. Although the Swin Transformer achieved suboptimal segmentation results, it performs poorly in small object segmentation and suffers from classification errors, as demonstrated in the visualizations of the second and third experimental groups in Fig.~\ref{fig3}. The U-Net network can correctly perform segmentation from a visual perspective, but it exhibits a noticeable jagged effect and segmentation gaps along the object edges.
%FAMSeg模型有效完成形态大小不同和方向不同的胎儿骨骼和颅脑的超声影像分割。虽然
\subsection{Ablation Study of the Model}
To further validate the reliability and feasibility of the FAMSeg design approach, we conducted an ablation study by controlling for identical variables, as shown in Table \ref{tab3}. Under the same conditions, we independently trained segmentation models by removing the Mamba structure and the feature fusion component, and compared the results with the baseline model. Experimental groups B and C demonstrate that integrating context fusion yields the most significant overall performance improvement. However, segmentation of small targets remains suboptimal. Therefore, as shown in experimental groups A and D, by combining both local and global information, we enhance the performance of small target segmentation.
Additionally, as shown in Table \ref{tab4}, we explored the impact of different optimizers and loss decay methods on model training performance and efficiency. While the combination of the AdamW optimizer and cosine loss decay yields results comparable to those of our proposed method, our design provides a better balance between overall segmentation and small target segmentation performance.
\begin{table*}
    \centering
    \setlength{\tabcolsep}{5pt}
    \caption{Comparison and validation experiments of different combinations of optimizers and loss reduction methods.}
    \begin{tabular}{cccccc|cccc}
    \hline
        SGD & Adam & AdamW & Step & Adadelta & COS & BG & FL & FB & mIoU$\uparrow$  \\ \hline
        $\checkmark$ & $\times$ & $\times$ & $\checkmark$ & $\times$ & $\times$ & 98.96 & 77.12 & 94.97 & 90.35  \\ \hline
        $\checkmark$ & $\times$ & $\times$ & $\times$ & $\checkmark$ & $\times$ & 99.04 & 78.98 & 95.34 & 91.12  \\ \hline
        $\checkmark$ & $\times$ & $\times$ & $\times$ & $\times$ & $\checkmark$ & 99.04 & 79.22 & 95.3 & 91.2  \\ \hline
        $\times$ & $\checkmark$ & $\times$ & $\times$ & $\times$ & $\checkmark$ & \textbf{99.08}  & 79.73  & 95.43  & \underline{91.41}   \\ \hline
        $\times$ & $\checkmark$ & $\times$ & $\times$ & $\checkmark$ & $\times$ & \underline{99.05} & 79.47 & 95.38 & 91.3  \\ \hline
        $\times$ & $\checkmark$ & $\times$ & $\checkmark$ & $\times$ & $\times$ & 99.01 & 77.5 & 95.26 & 90.59  \\ \hline
        $\times$ & $\times$ & $\checkmark$ & $\checkmark$ & $\times$ & $\times$ & 98.92 & 77.22 & 94.42 & 90.19  \\ \hline
        $\times$ & $\times$ & $\checkmark$ & $\times$ & $\times$ & $\checkmark$ & \textbf{99.08} & \underline{80.07} & \textbf{95.58} & \textbf{91.58}  \\ \hline
        $\times$ & $\times$ & $\checkmark$ & $\times$ & $\checkmark$ & $\times$ & \textbf{99.08}  & \textbf{80.21}  & \underline{95.44}  & \textbf{91.58}  \\ \hline
    \end{tabular}
    \label{tab3}
    \vspace{-5mm}
\end{table*}
\begin{table*}
    \vspace{-5mm}
    \centering
    \setlength{\tabcolsep}{13pt}
    \caption{Ablation study results of the core modules.}
    \begin{tabular}{l|cccc}
    \hline
        Item & BG & FL & FB & mIoU$\uparrow$  \\ \hline
        SegNeXt & 98.16  & 77.48  & 94.18  & 89.94   \\ \hline
        No mamba & \underline{99.05} & \underline{79.5} & \underline{95.32} & \underline{91.29}  \\ \hline
        Apomixis & 98.48 & 71.04 & 92.58 & 87.37  \\ \hline
        Ours & \textbf{99.08}  & \textbf{80.21}  & \textbf{95.44}  & \textbf{91.58}  \\ \hline
    \end{tabular}
    \label{tab4}
    \vspace{-6mm}
\end{table*}
\begin{figure}[htbp]
    \centerline{\includegraphics[width=\textwidth]{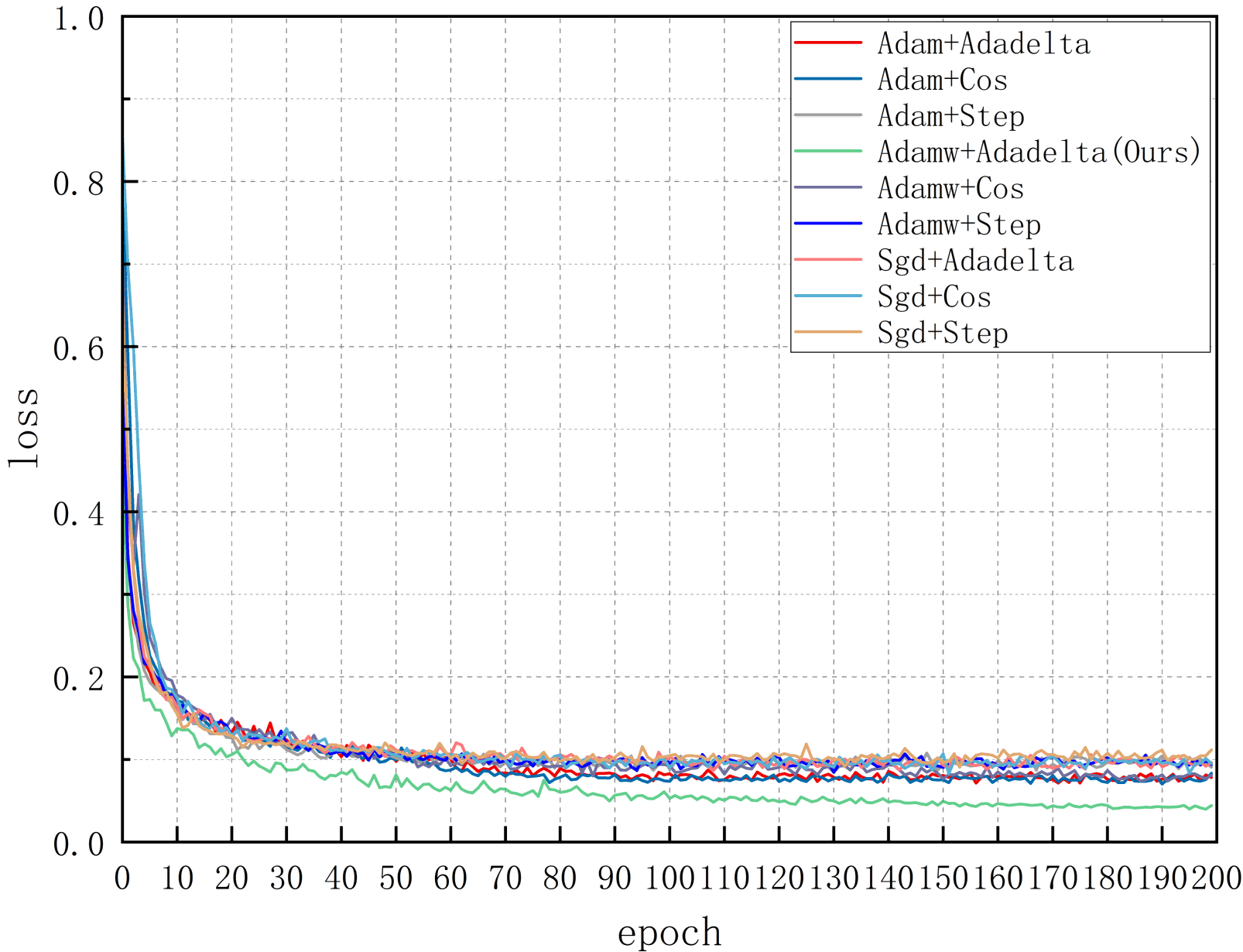}}
    \caption{Comparison experiment curve of different combinations of optimizers and loss reduction methods.}
    \label{fig4}
    \vspace{-8.6mm}
\end{figure}

Fig.~\ref{fig4} shows the training loss curves for different combinations of optimizers and decay methods, illustrating the convergence rate of the model. The AdamW and Adadelta combined optimization method used in this study achieved the fastest and most stable convergence.
\section{Conclusion}
In this paper, the FAMSeg semantic segmentation model is proposed to address the challenges of insufficient accuracy in small object segmentation and segmentation jagginess. The model explores the impact of lateral and vertical independent perspective convolution blocks, adaptive perception mechanisms, multi-scale contextual feature fusion, and different optimizers on segmentation performance. The proposed lateral and vertical independent perspective convolution blocks are designed to preserve the texture features of object contours. In addition, the adaptive perception mechanism enhances the reliance on both local features and global information, strengthening the model's ability to recover feature mappings. Although FAMSeg achieves significant results in fetal femur and brain segmentation, it still has some limitations. First, it cannot balance the accuracy of multi-object segmentation. Second, it is unable to align information from multiple models. Finally, the model training process is relatively time-consuming. In the future, we plan to further balance multi-object segmentation performance by combining large models with knowledge distillation techniques, and expand the application to a broader range of medical fields.
\vspace{-7mm}
% \subsection{Ablation Study of the Optimizer}
\section*{Acknowledgements}
This research was partially supported by the Natural Science Foundation of Guangxi under Grant No. 2020JJA170007, the Guangxi Natural Science Foundation under Grant No. 2024JJA141093, the National Natural Science Foundation of China under Grant No. 62162054,  the Key Research Project of Wuzhou University under Grant No. 2023C004 and No. 2024QN001, the Wuzhou Science and Technology Plan Project  under Grant No. 202302036.
\bibliographystyle{spmpsci_unsrt}
\bibliography{reference}

\begin{thebibliography}{10}
\providecommand{\url}[1]{{#1}}
\providecommand{\urlprefix}{URL }
\expandafter\ifx\csname urlstyle\endcsname\relax
  \providecommand{\doi}[1]{DOI~\discretionary{}{}{}#1}\else
  \providecommand{\doi}{DOI~\discretionary{}{}{}\begingroup \urlstyle{rm}\Url}\fi

\bibitem{pu2021automatic}
Pu, B., Li, K., Li, S., et~al.: Automatic fetal ultrasound standard plane recognition based on deep learning and iiot.
\newblock IEEE Transactions on Industrial Informatics \textbf{17}(11), 7771--7780 (2021)

\bibitem{gao2024graph}
Gao, Z., Tan, G., Wang, C., et~al.: Graph-enhanced ensembles of multi-scale structure perception deep architecture for fetal ultrasound plane recognition.
\newblock Engineering Applications of Artificial Intelligence \textbf{136}, 108,885 (2024)

\bibitem{pu2024unsupervised}
Pu, B., Lv, X., Yang, J., et~al.: Unsupervised domain adaptation for anatomical structure detection in ultrasound images.
\newblock In: Forty-first International Conference on Machine Learning (2024)

\bibitem{zhao2024farn}
Zhao, L., Tan, G., Wu, Q., et~al.: Farn: fetal anatomy reasoning network for detection with global context semantic and local topology relationship.
\newblock IEEE Journal of Biomedical and Health Informatics  (2024)

\bibitem{zhao2022ultrasound}
Zhao, L., Li, K., Pu, B., et~al.: An ultrasound standard plane detection model of fetal head based on multi-task learning and hybrid knowledge graph.
\newblock Future Generation Computer Systems \textbf{135}, 234--243 (2022)

\bibitem{pu2024m3}
Pu, B., Wang, L., Yang, J., et~al.: M3-uda: a new benchmark for unsupervised domain adaptive fetal cardiac structure detection.
\newblock In: Proceedings of the IEEE/CVF Conference on Computer Vision and Pattern Recognition, pp. 11,621--11,630 (2024)

\bibitem{yang2025bidirectional}
Yang, J., Lin, Y., Pu, B., Li, X.: Bidirectional recurrence for cardiac motion tracking with gaussian process latent coding.
\newblock Advances in Neural Information Processing Systems \textbf{37}, 34,800--34,823 (2025)

\bibitem{CHEN2024107898}
Chen, G., Tan, G., Duan, M., et~al.: Mlmseg: A multi-view learning model for ultrasound thyroid nodule segmentation.
\newblock Computers in Biology and Medicine \textbf{169}, 107,898 (2024)

\bibitem{9950712}
Lu, Y., Li, K., Pu, B., et~al.: A yolox-based deep instance segmentation neural network for cardiac anatomical structures in fetal ultrasound images.
\newblock IEEE/ACM Transactions on Computational Biology and Bioinformatics \textbf{21}(4), 1007--1018 (2024)

\bibitem{pu2022mobileunet}
Pu, B., Lu, Y., Chen, J., et~al.: Mobileunet-fpn: A semantic segmentation model for fetal ultrasound four-chamber segmentation in edge computing environments.
\newblock IEEE Journal of Biomedical and Health Informatics \textbf{26}(11), 5540--5550 (2022)

\bibitem{chen2024sam2}
Chen, T., Lu, A., Zhu, L., et~al.: Sam2-adapter: Evaluating \& adapting segment anything 2 in downstream tasks: Camouflage, shadow, medical image segmentation, and more.
\newblock arXiv preprint arXiv:2408.04579  (2024)

\bibitem{xing2024segmamba}
Xing, Z., Ye, T., Yang, Y., et~al.: Segmamba: Long-range sequential modeling mamba for 3d medical image segmentation.
\newblock In: International Conference on Medical Image Computing and Computer-Assisted Intervention, pp. 578--588. Springer (2024)

\bibitem{mamba2}
Dao, T., Gu, A.: Transformers are {SSM}s: Generalized models and efficient algorithms through structured state space duality.
\newblock In: International Conference on Machine Learning (ICML) (2024)

\bibitem{segnext}
Guo, M.H., Lu, C.Z., Hou, Q., et~al.: Segnext: Rethinking convolutional attention design for semantic segmentation.
\newblock Advances in neural information processing systems \textbf{35}, 1140--1156 (2022)

\bibitem{SGD}
Ketkar, N.: Stochastic gradient descent.
\newblock In: Deep learning with Python: A hands-on introduction, pp. 113--132. Springer (2017)

\bibitem{adamw}
Loshchilov, I., Hutter, F.: Decoupled weight decay regularization.
\newblock arXiv preprint arXiv:1711.05101  (2017)

\bibitem{adam}
Loshchilov, I., Hutter, F., et~al.: Fixing weight decay regularization in adam.
\newblock arXiv preprint arXiv:1711.05101 \textbf{5}, 5 (2017)

\bibitem{U-Net}
Xiao, T., Liu, Y., Zhou, B., et~al.: Unified perceptual parsing for scene understanding.
\newblock In: Proceedings of the European Conference on Computer Vision (ECCV), pp. 418--434 (2018)

\bibitem{chu2021twins}
Chu, X., Tian, Z., Wang, Y., et~al.: Twins: Revisiting spatial attention design in vision transformers.
\newblock arXiv preprint arXiv:2104.13840  (2021)

\bibitem{liu2021Swin}
Liu, Z., Lin, Y., Cao, Y., et~al.: Swin transformer: Hierarchical vision transformer using shifted windows.
\newblock arXiv preprint arXiv:2103.14030  (2021)

\bibitem{Howard_2019_ICCV}
Howard, A., Sandler, M., et~al.: Searching for mobilenetv3.
\newblock In: The IEEE International Conference on Computer Vision (ICCV), pp. 1314--1324 (2019).
\newblock \doi{10.1109/ICCV.2019.00140}

\bibitem{liu2022convnet}
Liu, Z., Mao, H., Wu, C.Y., et~al.: A convnet for the 2020s.
\newblock Proceedings of the IEEE/CVF Conference on Computer Vision and Pattern Recognition (CVPR)  (2022)

\bibitem{yu2021bisenet}
Yu, C., Gao, C., Wang, J., et~al.: Bisenet v2: Bilateral network with guided aggregation for real-time semantic segmentation.
\newblock International Journal of Computer Vision pp. 1--18 (2021)

\end{thebibliography}

\end{document}